\documentclass[conference]{IEEEtran}
\IEEEoverridecommandlockouts
\usepackage{cite}
\usepackage{amsmath,amssymb,amsfonts}
\usepackage{algorithm}
\usepackage[noend]{algpseudocode}
\usepackage{graphicx}
\usepackage{textcomp}
\usepackage{xcolor}
\usepackage{subcaption}
\usepackage{url}
\usepackage{booktabs}
\usepackage{multirow}
\usepackage{makecell}
\usepackage{flushend}
\usepackage{pifont}
\newcommand{\cmark}{\ding{51}}%
\newcommand{\xmark}{\ding{55}}%

\def\BibTeX{{\rm B\kern-.05em{\sc i\kern-.025em b}\kern-.08em
    T\kern-.1667em\lower.7ex\hbox{E}\kern-.125emX}}
\begin{document}

\title{Mixed-TD: Efficient Neural Network Accelerator with Layer-Specific Tensor Decomposition}
\author{\IEEEauthorblockN{Zhewen Yu, Christos-Savvas Bouganis}
\textit{Imperial College London}\\
London, UK \\
\{zhewen.yu18, christos-savvas.bouganis\}@imperial.ac.uk}
\maketitle

\begin{abstract}
Neural Network designs are quite diverse, from VGG-style to ResNet-style, and from Convolutional Neural Networks to Transformers. Towards the design of efficient accelerators, many works have adopted a dataflow-based, inter-layer pipelined architecture, with a customized hardware towards each layer, achieving ultra high throughput and low latency. The deployment of neural networks to such dataflow architecture accelerators is usually hindered by the available on-chip memory as it is desirable to preload the weights of neural networks on-chip to maximise the system performance. To address this, networks are usually compressed before the deployment through methods such as pruning, quantization and tensor decomposition. In this paper, a framework for mapping CNNs onto FPGAs based on a novel tensor decomposition method called Mixed-TD is proposed. The proposed method applies layer-specific Singular Value Decomposition (SVD) and Canonical Polyadic Decomposition (CPD) in a mixed manner, achieving 1.73$\times$ to 10.29$\times$ throughput per DSP to state-of-the-art CNNs. Our work is open-sourced: \url{https://github.com/Yu-Zhewen/Mixed-TD}.

\end{abstract}


\section{Introduction}
The recent advances in Machine Learning (ML) research have led to the design of a large and diverse set of neural network designs. At a macroscopic level, popular designs include Convolutional Neural Networks (CNNs) and Transformers, where at a microscopic level the above structures contain layers with different properties, including number of channels, kernel size, residual connection, etc. Towards the design of neural network accelerators, many works adopted a dataflow architecture \cite{venieris2018toolflows}, customizing the computational pipeline to each layer to maximise efficiency and achieve high throughput. 

A challenge for the dataflow-based accelerators is the storing of the parameters of the neural network to on-chip memory. Let's consider the AMD Alveo U250, a data-center acceleration card. Despite its high performance, this card has a limited internal SRAM capacity of only 54MB. A ResNet-50, a popular neural network architecture with 23 million parameters, requires a storage capacity of approximately 92MB in floating-point format, which exceeds the SRAM capacity. Apart from the memory capacity, the on-chip memory bandwidth also becomes a limitation when the architecture requires access to a large number of the parameters concurrently to support parallel computation for enhanced performance. Storing the parameters to off-chip memory addresses the capacity problem but penalises the performance of the system.

To reduce the memory footprint of the model, existing approaches compress the weights of a pre-trained neural network and fine-tune the compressed weights before the network is mapped to a dataflow-based accelerator. In the case of a pre-trained deep neural network, the data distribution and error tolerance vary across different parts of the network \cite{wang2019deep}. As such, it is necessary to make the compression method fine-grained and layer-specific to avoid significant accuracy degradation. In previous work on weights pruning, unstructured ways of pruning bring a larger compression ratio with negligible accuracy degradation compared with the channel-wise structured pruning \cite{liu2021leveraging, meng2021fixyfpga}. Similarly, in weights quantization, mixed precision and block floating point methods are favoured compared with uniform wordlength and range \cite{rasoulinezhad2019pir, alonso2021elastic, latotzke2022design}.

In this paper, we explore a different dimension for performing fine-grained, layer-specific, and hardware-friendly weights compression, by using Tensor Decomposition techniques. Furthermore, we propose the use of an ML-based proxy for predicting the performance of the compressed model during its mapping to an FPGA, making it possible to explore the large design space defined by the introduced Tensor decomposition schemes. Our performance evaluation shows that the proposed approach can achieve significant weight compression with negligible accuracy penalty, as well as result in designs with competitive latency and throughput to state-of-the-art approaches. 
\begin{figure}[t]
    \centering
    \includegraphics[width=0.485\textwidth]{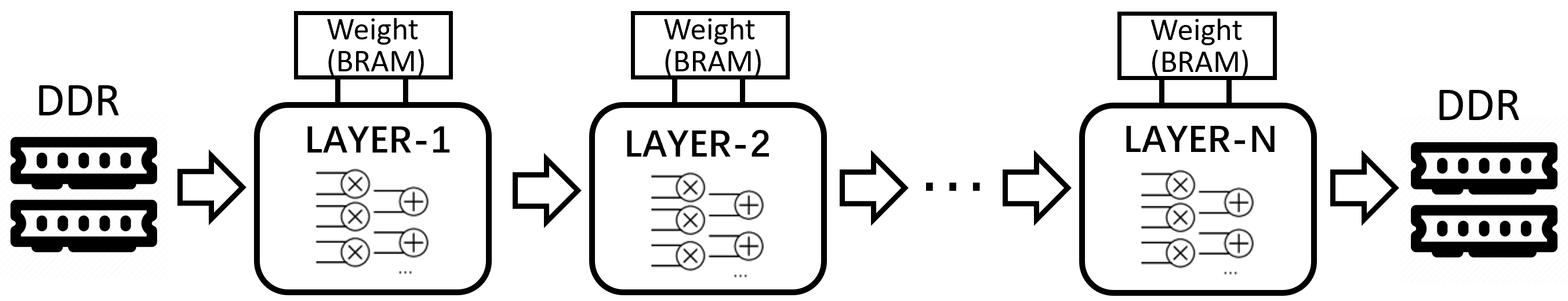}
    \caption{An example of a dataflow architecture, where each layer of the network has its own customized hardware. Inter-layer and intra-layer pipelines are usually applied.}
    \label{fig:dataflow}
\end{figure}

The key contributions of this paper are as follows:
\begin{itemize}
    \item A novel weight compression method, Mixed-TD, that is based on tensor decomposition techniques. It extends current approaches by introducing for the first time a layer-specific mixture of Singular Value Decomposition (SVD) and Canonical Polyadic Decomposition (CPD). 
    \item An efficient method for fast design space exploration through the evolutionary search and a random-forest-based throughput predictor.
    \item A dataflow-based accelerator design achieving low latency, high throughput, and negligible accuracy loss at the same time. In terms of the \textit{Throughput per DSP}, we are achieving gains of 1.73$\times$ to 10.29$\times$ compared to existing work. 
\end{itemize}

\section{Background}
\subsection{Tensor Decomposition}
Tensor decomposition expresses one tensor as a set of elementary and simpler tensors which act on each other \cite{heath1986computing, kolda2009tensor}. The decomposition of a tensor $\mathcal{A}$ can be derived through the following optimization problem:

\begin{equation}
\begin{aligned}
    \min_{a_{1}, a_{2} \dots a_{M}}~r_{1}, r_{2} \dots r_{M} \\
    s.t.~||\mathcal{A}-f(a_{1}, a_{2} \dots a_{M})|| \leq \epsilon,
\end{aligned}
\end{equation}
where $a_{i}, i\in[1,M]$ is the set of elementary tensors and $f$ is the approximation function $f$ so that the given $M_{th}$-order tensor $\mathcal{A}$ is approximated within the desired error boundary $\epsilon$ and the ranks of these elementary tensors $r_{i}$ are also minimised. Depending on the operations inside the function $f$, there are different formats of tensor decomposition available. SVD and CPD are two of them that have been well studied in compressing the weights of neural networks \cite{kossaifi2016tensorly}.

SVD targets the compression of a given $2_{nd}$-order tensor, which is $M=2$ and $\mathcal{A} \in \mathbb{R}^{d_1 \times d_2}$. $\mathcal{A}$ is decomposed into the form of 
\begin{equation}
    \mathcal{A} \approx U_{r} \Sigma_{r} V^{T}_{r}, 
    \label{equ:svd}
\end{equation}
where the rows of $U_{r}$ and the columns of $V_{r}$ are orthogonal bases, and $\Sigma_{r}$ is a diagonal matrix containing the top-$r$ singular values in descending order. After absorbing the diagonal matrix $\Sigma_{r}$ into $U_{r}$ and $V_{r}$, the final format becomes the product of two tensors and the number of parameters remaining is $(d_1+d_2)r$. SVD can also be used to compress the tensor with higher-order ($M>2$) which requires reducing the order of the target tensor with slicing and reshaping operations beforehand \cite{yu2023svd}. 

\begin{figure}[t]
    \centering
    \hspace{1.1cm}
    \begin{subfigure}[c]{0.35\textwidth}
        \centering
        \includegraphics[width=\textwidth]{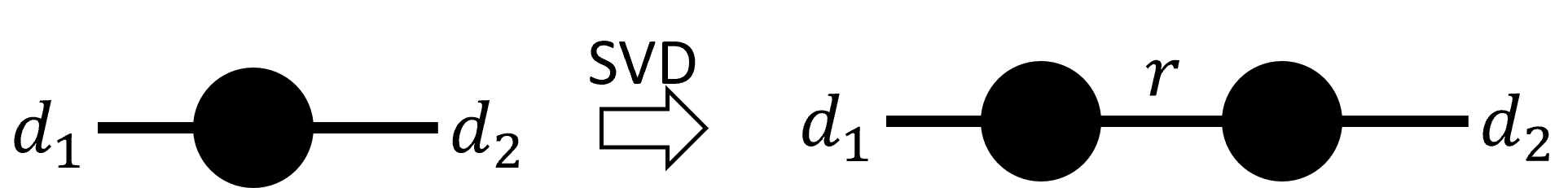}
    \end{subfigure} \hfill
        \vspace{4mm}
    \begin{subfigure}[c]{0.35\textwidth}
        \centering
        \includegraphics[width=\textwidth]{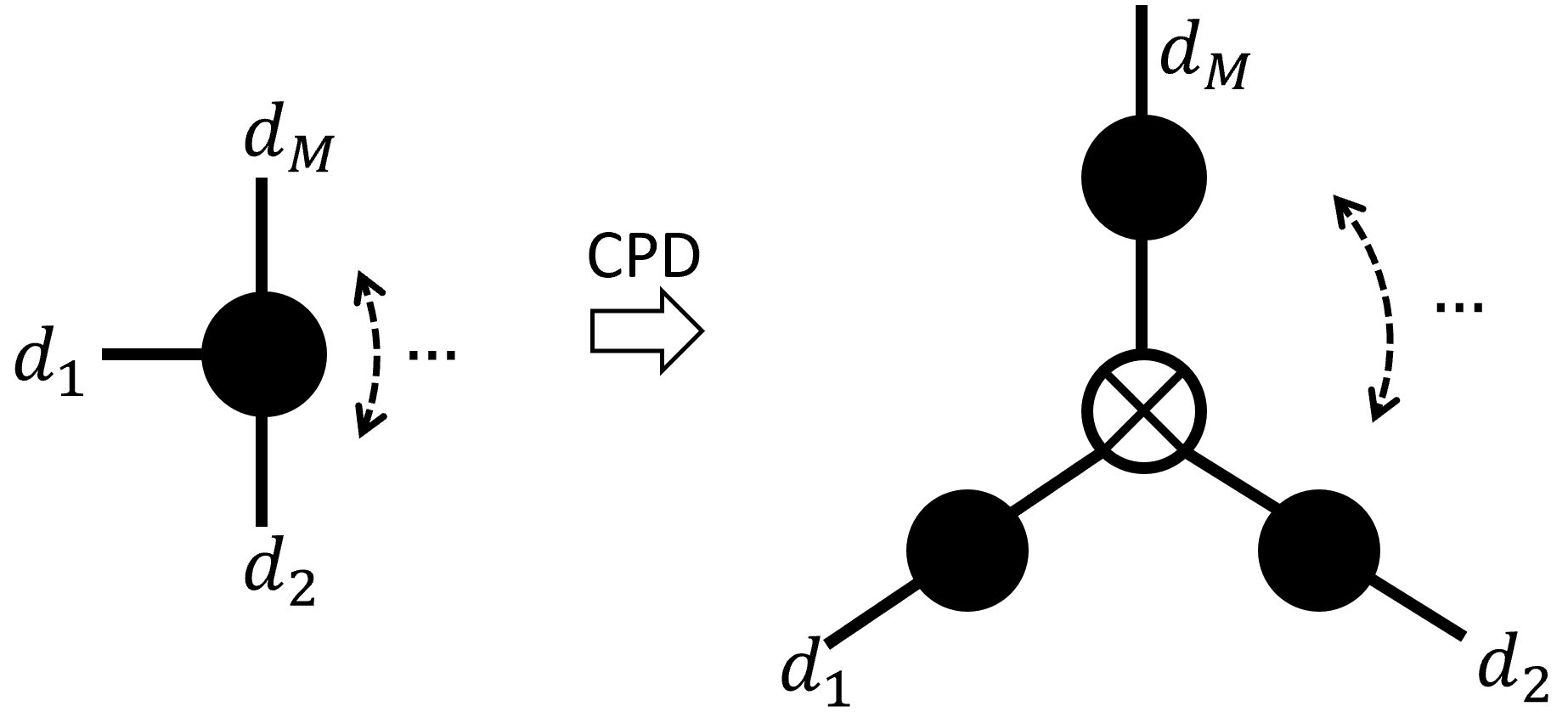}
    \end{subfigure}
    \caption{Represent SVD and CPD in the Tensor Diagram Notation\cite{tensor_network}. Each solid node denotes a $M_{th}$-order tensor with $M$ edges. Edges connected together represent the tensor contraction operation between two tensors. In CPD, the hollow node with a cross inside represents the sum of the outer product.}
    \label{fig:svd_cpd_tensor_diagram}
\end{figure}

CPD can be applied directly to a high-order tensor without the need of reducing its order to $M=2$ like SVD. Given the $M_{th}$-order tensor, $\mathcal{A} \in \mathbb{R}^{d_1 \times d_2 \times \dots d_{M}}$, its CPD format can be represented as the sum of the outer product of $1_{st}$-order tensors.
\begin{equation}
    \mathcal{A} \approx \sum_{i=1}^{r} a_{1,i} \otimes a_{2,i} \otimes \dots a_{M,i}
    \label{equ:cpd}
\end{equation}
This set of $1_{st}$-order tensors can be computed via the Alternating Least Squares (ALS) method \cite{kolda2009tensor}, and the number of parameters remaining is $(d_1 + d_2 + \dots d_{M})r$. The difference between SVD and CPD is visualised in Fig.~\ref{fig:svd_cpd_tensor_diagram}.

\subsection{Related Work}
Many state-of-the-art approaches that target the acceleration of CNNs are based on dataflow architectures, as they reduce the off-chip memory accesses, and customize the hardware pipeline to the properties of the targeted CNN load.

fpgaConvNet \cite{venieris2016fpgaconvnet} generates a highly optimized architecture based on the target CNN and FPGA board. They considered the splitting of the whole data flow into multiple partitions, where each partition only contains a subgraph of the neural network, allowing the tool to map large CNN models to small devices through reconfiguration. The tool performs a faithful mapping of the CNN to the FPGA, assuming an already optimized CNN model as its input.


FINN \cite{petrica2020memory} implemented the dataflow architecture through extreme network quantization. The authors fit a ResNet-50 network on a U250 device, after quantizing the weights and the activations of the network to 1 bit and 2 bits respectively, except the first and last layers which are quantized to 8 bits. Their mixed precision implementation achieves an impressive 2703 Frames Per Second (FPS) but they also reported 9.8 percentage points (pp) top-1 accuracy degradation on the ImageNet dataset compared to the floating point version of the network.

HPIPE \cite{hall2020hpipe} explored weights compression through unstructured sparsity. The authors eliminated $85\%$ of the weights from ResNet-50 and encoded the remaining parameters in a compressed format to save memory storage. The authors reported a 5.2pp accuracy degradation and an achieved throughput of 4550 FPS on a Stratix 10 2800 FPGA.

Our approach takes a step to a different direction from prior work as it explores a different dimension for performing fine-grained, layer-specific weights compression, which achieves competitive performance compared with existing methods including unstructured sparsity and mixed precision quantization.

\section{Mixed Tensor Decomposition}
\label{sec:mixed_td}
This section introduces our proposed fine-grained compression method, termed Mixed-TD, which opens the space for applying layer-specific SVD and CPD decompositions in a mixed manner.

\subsection{Compute Decomposed Convolution}
Let us consider an $N$-layer CNN whose weights are represented as $4_{th}$-order tensors $\mathcal{W}_{j} \in \mathbb{R}^{c_{j+1} \times c_{j} \times k_{j} \times k_{j}}, j \in [1,N]$, where $c_{j+1}$, $c_{j}$ and $k_{j}\times k_{j}$ are denoting the number of output channels, input channels and the kernel size of the $j_{th}$ convolutional layer respectively. Its input and output are denoted as $\mathcal{X}_{j} \in \mathbb{R}^{b \times m_{j} \times n_{j} \times c_{j}}$ and $\mathcal{X}_{j+1} \in \mathbb{R}^{b \times m_{j+1} \times n_{j+1} \times c_{j+1}}$, where $b$ denotes the batch size, $m_{j}$ and $n_{j}$ denote the spatial size of the feature map. 

The convolution operation can be represented as the tensor contraction along $c_{j}$ and $k_{j}$ dimensions,
\begin{equation}
    \mathcal{X}_{j+1} = \sum_{c_{j}, k_{j}}(\mathcal{W}_{j} \cdot S(\mathcal{X}_{j})),
\end{equation}

where the sliding window function $S$ performs the padding and striding, converting $\mathcal{X}_{j}$ into the shape of $b \times m_{j+1} \times n_{j+1} \times c_{j} \times k_{j} \times k_{j}$. 

Our proposed approach performs tensor decomposition on the weights tensor at design time. After decomposition, the tensor $\mathcal{W}_{j}$ is substituted by the form of \eqref{equ:svd} or \eqref{equ:cpd}, transforming as such the computation from one tensor contraction operation into multiple consecutive ones, but with a reduced total number of elements. 

\subsection{Layer-Specific Design Choices}
\label{subsec:layer_spe}
In the proposed Mixed-TD method, there are three types of layer-specific design choices.

\textbf{SVD or CPD:} For each convolutional layer, its 4-d weight tensor can be decomposed either in SVD or CPD format. For SVD, the 4-d tensor is reshaped into 2-d and then decomposed into the product of two 2-d tensors, as \eqref{equ:svd} shows. Otherwise, for CPD, the 4-d tensor is decomposed to the sum of the outer product of four 1-d tensors, as \eqref{equ:cpd} shows. The main difference between these two formats is the number of tensors left after decomposition as well as their representation abilities. In Section \ref{sec:arch} and \ref{sec:exp}, we will see the number of tensors left affects the accelerator performance in a dataflow architecture design. A rule of thumb is that the CPD format has more overhead than SVD in hardware design but in terms of the impact on network accuracy, whether to use SVD or CPD is really layer-specific. We use the enumeration variable $t_{j}=\{0:SVD,1:CPD\}$ to represent this decision.

\textbf{Channels grouping:} Instead of directly applying the tensor decomposition to the entire $\mathcal{W}_{j}$, we can choose to slice and split $\mathcal{W}_{j}$ into multiple chunks where each chuck can then be decomposed individually. As in CNN designs $c_{j+1}$ and $c_{j}$ are usually much larger than $k_{j}$, our method provides the options of slicing the first ($c_{j+1}$) and the second dimension ($c_{j}$) of $\mathcal{W}_{j}$ into $g_{1,j}$ and $g_{2,j}$ groups respectively. As such, for each $\mathcal{W}_{j}$, it can be split into $g_{1,j}g_{2,j}$ chunks in total, and each chuck has the shape of $\mathbb{R}^{\frac{c_{j+1}}{g_{1,j}} \times \frac{c_{j}}{g_{2,j}} \times k_{j} \times k_{j}}$. The intuition behind offering this type of design choice is to leverage the diversity and similarity of feature maps \cite{goetschalckx2018efficiently}.

\textbf{Rank selection:} In both SVD and CPD methods, a key design choice is the rank selection, as it controls the compression ratio of the decomposition. Within a network, the redundancy of each layer varies considerably and setting a uniform compression ratio dramatically hurts the accuracy \cite{yang2020automatic}. Therefore, it is critical to have the layer-specific choice of the rank $r_{j}$ after searching for the optimal combination of them.
\begin{figure*}[h]
     \centering
    \begin{subfigure}[c]{0.54\textwidth}
        \centering
        \includegraphics[width=\textwidth]{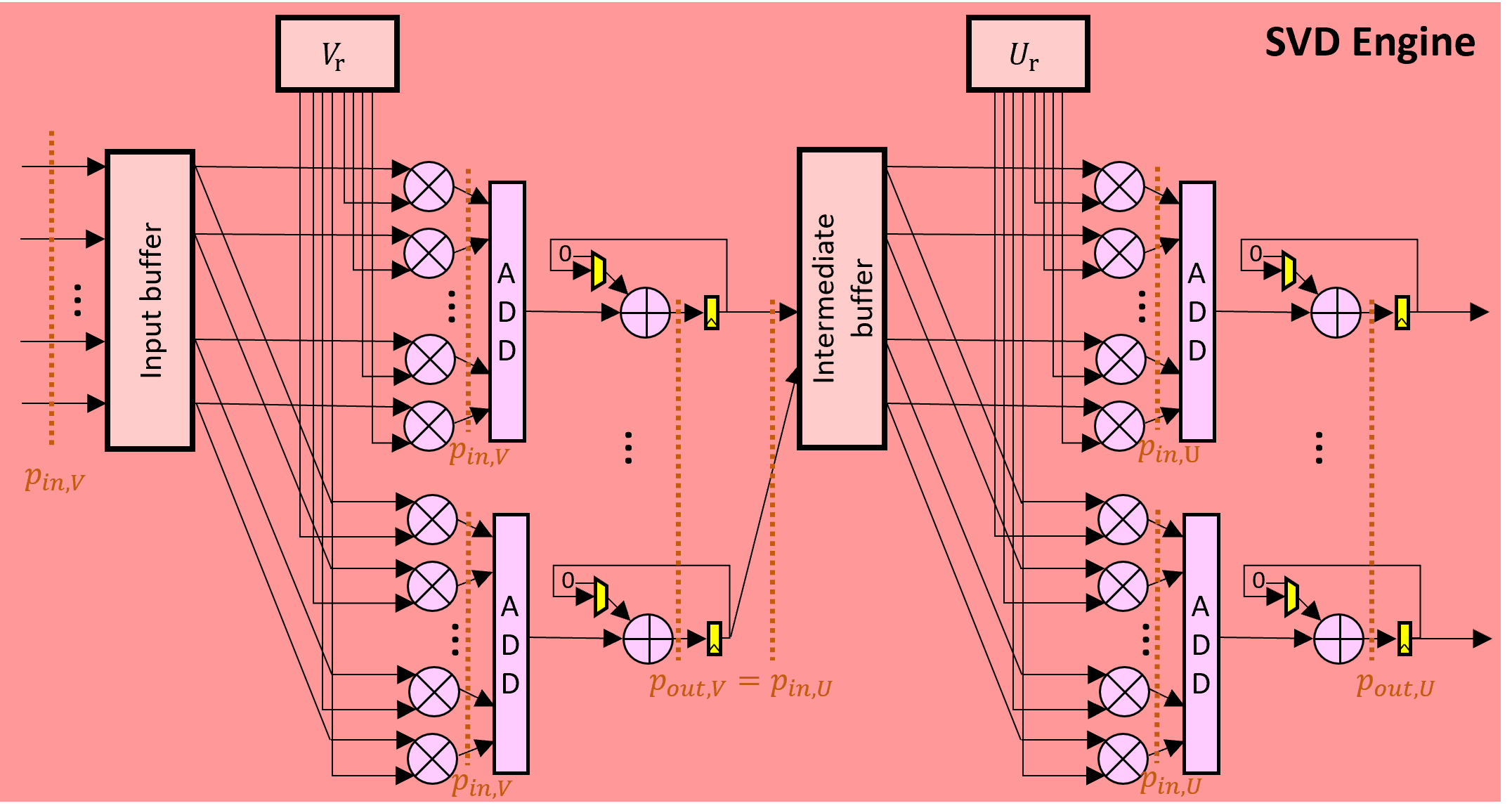}
    \end{subfigure}
    \begin{subfigure}[c]{0.42\textwidth}
        \centering
        \footnotesize
        \begin{tabular}{@{}ccccc@{}}
        \toprule
                     & SVD Engine & CPD Engine \\
        \midrule
             Rank    & \multicolumn{2}{c}{$r$} \\
        \midrule
             Inner Product & 2 & 2 \\
             Outer Product & 0 & 2 \\
             Weights & $U_{r}$, $V_{r}$ & $a_{\{1,2,3,4\},r}$ \\
             \multirow{2}*{Unroll factors} & $p_{in,U}$, $p_{in,V}$ & $p_{in,\{1,2,3,4\}}$ \\
             ~ & $p_{out,U}$, $p_{out,V}$ & $p_{out,\{1,2,3,4\}}$\\
        \bottomrule
        \end{tabular}
    \end{subfigure}
    
    \hfill
    
    \begin{subfigure}[c]{0.96\textwidth}
        \centering
        \includegraphics[width=\textwidth]{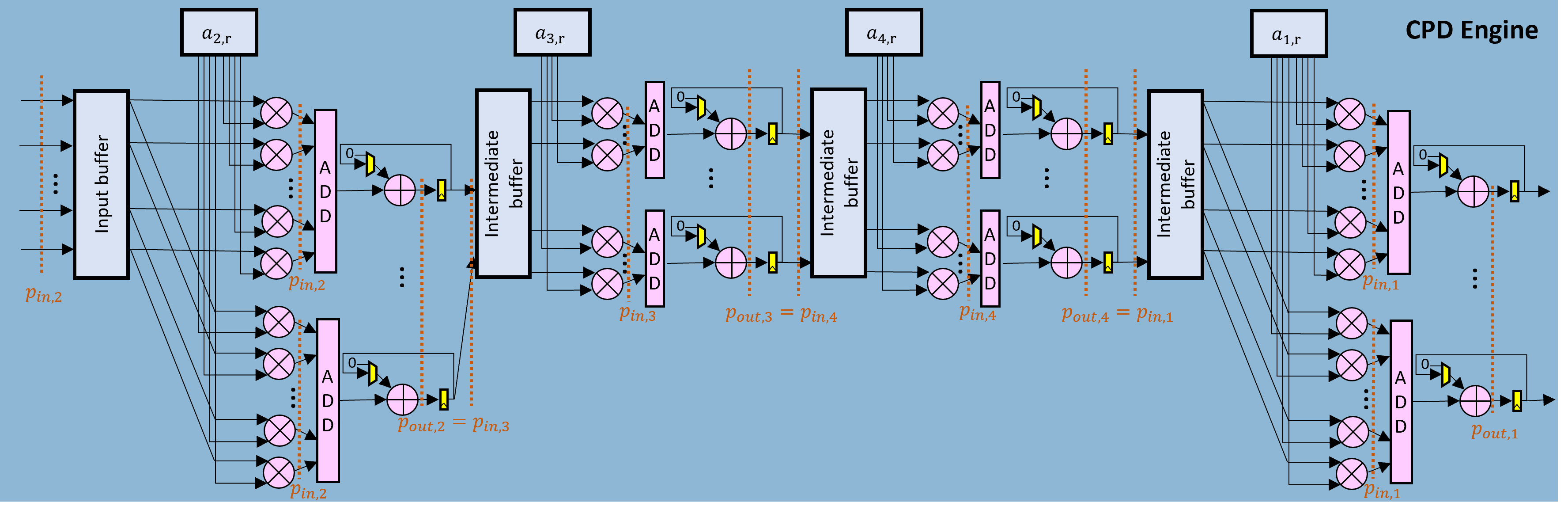}
    \end{subfigure} 
    \caption{Internal architecture of the proposed SVD Engine and CPD Engine. Both architectures implement the convolution on the decomposed weight tensors, but they differ in the number of stages of computation as well as the dataflow between stages.}
    \label{fig:engine}
\end{figure*}

\section{Accelerator Design}
\label{sec:arch}
This section introduces the accelerator designs for deploying networks compressed by our Mixed-TD method. A high level description of the dataflow architecture is given in section \ref{subsec:dataflow}, the internal structures of compute engines are discussed in section \ref{subsec:engine}, and the data connections between engines are elaborated in section \ref{subsec:cross_layer}.

\subsection{Dataflow Architecture}
\label{subsec:dataflow}
We extend an open-source dataflow architecture accelerator fpgaConvNet \cite{venieris2016fpgaconvnet, montgomerie2019power}. As we are targeting both high throughput and low latency, we did not use the device reconfiguration technique in our work. Each layer, including convolution, relu, pooling, elementwise addition, fully connected, and etc., is mapped to a dedicated computation engine. These engines are connected in a pipeline manner to maximise overall throughput. During inference, the input data is sent from DDR to FPGA using the AXI-Stream interface, propagating through all computation engines, and the final prediction results are sent back to DDR using AXI-Stream. The system processes input data in batches and the pipeline is emptied between different batches.

\subsection{SVD v.s. CPD Engine}
\label{subsec:engine}
Each compressed convolutional layer of the network is mapped to either the SVD Engine or the CPD Engine with all the weights stored at the on-chip memories and preloaded before the inference starts. Both the SVD Engine and CPD Engine adopt the structure of input buffer, MAC units and accumulation module to compute the tensor contractions (Fig.~\ref{fig:engine}).

\begin{itemize}
    \item Input Buffer: It applies the sliding window function to input feature maps using a set of $p_{in}$ line buffers. At every clock cycle, each line buffer fetches one word of data from the previous layer and dispatches a window with the size of $k_{j} \times k_{j}$. 
    \item MAC Units: $p_{out}$ MAC units operate in parallel. Each MAC unit, which contains a multiplier array followed by an adder tree, is responsible for a vector-vector multiplication. The MAC units are fed by the input buffer as well as the preloaded weights.
    \item Accumulation Module: It gathers and accumulates the partial sums produced by the MAC units before dispatching the data into the next stage of computation.
\end{itemize}

Now, we highlight the differences between the proposed SVD Engine and CPD Engine.
\begin{itemize}
    \item In the case of SVD decomposition, the convolution kernel is decomposed into \textbf{two} stages of tensor contraction, as \eqref{equ:svd} shows. In the case of CPD decomposition, the computation is decomposed to \textbf{four} stages instead, as \eqref{equ:cpd} shows. In each stage, $p_{in}$ and $p_{out}$ can be tuned individually to trade resources for throughput.
    \item In the SVD Engine, the buffer \textbf{broadcasts} all the inputs to $p_{out}$ of MAC units concurrently to compute the inner products. On the contrary, in the CPD Engine, at the second and the third stages of computation (involving $a_{3,r}$ and $a_{4,r}$), the data coming out of the buffer is \textbf{scattered} to the $p_{out}$ MAC units instead to compute the outer products.
\end{itemize}

\subsection{Cross-layer Data Flow}
\label{subsec:cross_layer}
In the case of decomposed convolutions without channel grouping (as explained in Section \ref{sec:mixed_td}), each layer will require its own engine, and data will flow between layers in the form of a flattened 4-dimensional tensor, in the form of $b \times m_{j} \times n_{j} \times c_{j}$, where $j \in [1,N]$. In this notation, $b$ represents the batch size, $m_{j}$ and $n_{j}$ represent the spatial size of the feature map, and $c_{j}$ represents the number of feature maps in the $j$-th layer. 

In the case where channel grouping is applied, i.e. either $g_{1,j}$ or $g_{2,j}$ is greater than 1, there will be $g_{1,j}g_{2,j}$ engines per layer. In this case, data flows into the engines in a flattened form, as $b \times m_{j} \times n_{j} \times \frac{c_{j}}{g_{2,j}} \times g_{2,j}$. After splitting of the data into $g_{2,j}$ groups is performed, each engine fetches data in the form of $b \times m_{j} \times n_{j} \times \frac{c_{j}}{g_{2,j}}$. Similarly, at the layer's output, data is in the form of $b \times m_{j+1} \times n_{j+1} \times \frac{c_{j+1}}{g_{1,j}} \times g_{1,j}$.

In addition, if $g_{1,j} \neq g_{2,j+1}$, the data must be rearranged from $\frac{c_{j+1}}{g_{1,j}} \times g_{1,j}$ to $\frac{c_{j+1}}{g_{2,j+1}} \times g_{2,j+1}$ before being fed into the engines of the next layer. This is achieved using an array of FIFOs with a width set to the Least Common Multiple (LCM) of $g_{1,j}$ and $g_{2,j+1}$. The data streams are written and read from this FIFO array in a round-robin fashion to effectively rearrange the data.

\begin{figure*}[t]
    \centering
    \includegraphics[width=0.99\textwidth]{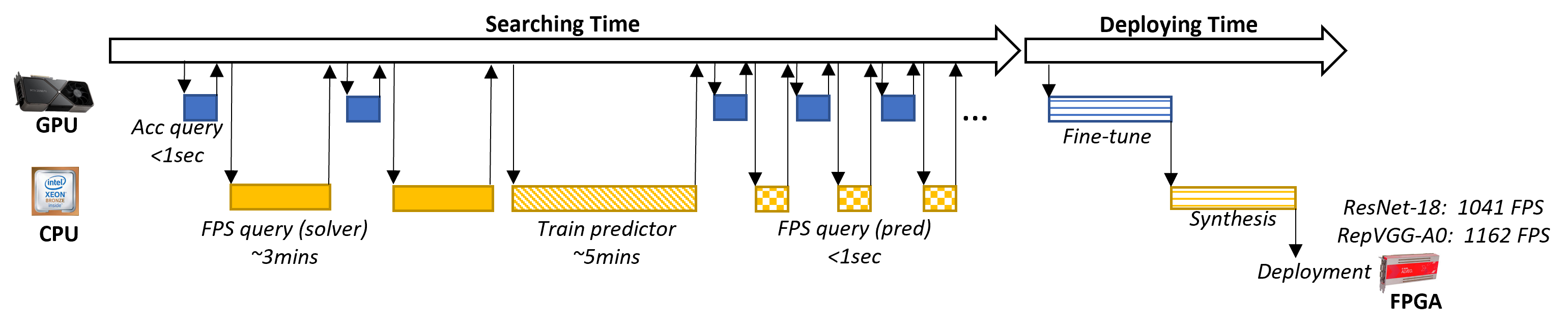}
    \caption{Design flow of our system. The flow includes the \textit{search} and the \textit{deployment} stages. During the \textit{search} stage, we query the accuracy and the throughput of each design point and perform the design space exploration to identify the optimal design. The optimal design is then fine-tuned to improve its accuracy before being synthesised and deployed on the FPGA device.}
    \label{fig:dse_flow}
\end{figure*}

\section{Design Space Exploration}
Having introduced both the compression algorithm and accelerator architecture, this section focuses on their integration, and how we can efficiently identify the optimal design point.

\subsection{Problem Formulation}
We define the optimal design point as the one that maximizes the prediction accuracy of the network, while satisfying the given resource budget and the target throughput in Frames Per Second (FPS). Several variables affect the design point, including the applied decomposition parameters $\tau=\{g_{1,j},g_{2,j},t_{j},r_{j}\}, j\in[1,N]$ (defined in section \ref{subsec:layer_spe}); as well as the accelerator's unrolling factors $\psi=\{p_{in,j}, p_{out,j}\}, j\in[1,N]$ (defined in section \ref{subsec:engine}).

\begin{equation}
    \max_{\tau, \psi} ACC~~~s.t.~FPS \geq FPS_{target},~~ RSC \leq RSC_{budget}
\end{equation}

The accuracy of the network depends on the choice of $\tau$ only. The FPS depends on both $\tau$ and $\psi$, as $\tau$ controls the per-layer workload and $\psi$ determines the initiation interval of the computation loop. The total resource is represented by the sum of per-layer resources, depending on $\psi$ only. To efficiently solve the constrained optimization problem, we decouple the searches of decomposition parameters $\tau$ and unrolling factors $\psi$, which are elaborated in the following two sections respectively.

\subsection{Evolutionary Search}
Due to the fine-grained and layer-specific decisions made by the proposed Mixed-TD method, the design space defined by $\tau$ is extremely large. To illustrate, ResNet-18 alone consists of $5.7\times10^{29}$ possible candidate designs.

As such, for the efficient search of $\tau$, we adopt the evolutionary searching algorithm proposed in \cite{dai2019chamnet}. Initially, we randomly sample from the design space and keep only the valid design points that satisfy the throughput and resource constraints until we obtain $|\textbf{P}|$ valid design points. These $|\textbf{P}|$ design points form the first generation of the ``population", referred to as ``parents". Using mutation and crossover, we generate $|\textbf{C}|$ new valid samples, referred to as ``children". The parents and children are then ranked together based on their prediction accuracy, and only the top-$|\textbf{P}|$ samples are retained to become the parents of the next generation. This ensures the continuation of high-performing design points throughout the evolution process.

\begin{algorithm}[h]
\caption{Constrained Evolutionary Search}
\begin{algorithmic}[1]
\Procedure{Validate}{sample $i$}
\State query accuracy of $i$
\State query throughput of $i$
\Return $FPS \geq FPS_{target}$ 
\EndProcedure

\Procedure{Searching}{}
    \State random sample $|\textbf{P}|$ valid designs
    \While {step $\leq$ max\_steps}
        \State mutate \textbf{P}, obtain $|\textbf{C}|/2$ new valid samples. 
        \State crossover \textbf{P}, obtain $|\textbf{C}|/2$ new valid samples. 
        \State sort $\textbf{P} \cup \textbf{C}$ by accuracy
        \State keep top-$|\textbf{P}|$ samples
    \EndWhile
\EndProcedure
\end{algorithmic}
\end{algorithm}

\subsection{Performance Predictor}
To further speed up the search of the evolutionary algorithm, it is crucial to minimize the time spent on evaluating the accuracy and throughput of each design point. 

For accuracy estimation, we use the decomposed network without fine-tuning and evaluate it on a single batch of data, rather than the entire validation dataset. This significantly reduces the time required for accuracy queries, reducing the duration from minutes to just seconds on a desktop GPU.

For throughput estimation, it is necessary to solve the following resource allocation problem that identifies the optimal configuration of unrolling factors $\psi$.
\begin{equation}
    \max_{\psi} FPS~~~s.t.~RSC \leq RSC_{budget}
\end{equation}
The unrolling factors impact both the resource utilization and system throughput. The optimal configuration is which maximizes the system throughput by balancing the delays of pipeline stages , under a given resource constraint. 

Existing optimizers are based on heuristics and take minutes or hours to run for networks with about 10 to 50 layers \cite{montgomerie2022samo}. Moreover, such a optimization process has to be repeated whenever the compression decisions change, limiting the search speed of Algorithm 1 and prohibiting the application to our case.

To address this challenge, the paper proposes building a proxy that predicts the achievable throughput for a specific configuration based on network characteristics and the optimizer used. For this purpose, a random forest was selected as the proxy model.

Our approach is as follows: In the first few generations of the evolutionary search, we use the solver from \cite{montgomerie2022samo} to obtain the throughput of each design, which we save and use to build a dataset. At a predetermined point in the search, we use this dataset to train a random forest regressor, which is then used to predict the throughput for subsequent designs during the evolutionary algorithm. Training the regressor takes only a few minutes on a desktop CPU, and its inference time is less than a second.

Our work differs from previous study \cite{dudziak2020brp}, which aimed to build performance predictors for general-purpose computing architectures, such as CPUs, GPUs, or systolic array accelerators, where the architecture is fixed and only the workload changes during the search. In contrast, our work targets dataflow architectures, where the hardware is changed and customized for each design point.

\begin{table*}[t]
\setlength\tabcolsep{5.5pt}
\centering
\caption{Performance and resource comparison with existing work on CNN-FPGA accelerators. In terms of the compression method, ``Q" stands for quantization, ``S" stands for sparsity (weights pruning) and ``TD" stands for tensor decomposition. In terms of performance metrics, both FPS for batch size 1, and peak FPS for a large batch size (we and \cite{yu2021streamsvd} use 256, \cite{petrica2020memory} uses 1024, and others did not provide the details) are reported. As the pipeline is emptied between different batches, the pipeline depth impacts differently those two FPS metrics. The resources of Intel devices have already been converted to the equivalent resources on AMD devices, where 1 Intel ALM = 1.8 AMD LUT \cite{shannon2015technology}, and 1 Intel DSP = 2 AMD DSP \cite{hall2020hpipe}.}
 \begin{tabular}{@{}cccccccccccccccc@{}}
\toprule
& Dataflow & Network & Method & Acc. & Device &\makecell{Freq.\\(MHz)} & URAM & BRAM & kLUT & DSP & \makecell{FPS \\ \textit{batch 1}} & \makecell{FPS \\ \textit{peak}} \\
\midrule
N3H-Core \cite{gong2022n3h} & \xmark & ResNet-18 & Q & 70.4 & XC7Z045 & 100 & - & 541 & 153 & 900 & 31 & 123  \\
FILM-QNN \cite{sun2022film} & \xmark & ResNet-18 & Q & 70.5 & ZCU102  & 150 & - & 881 & 180 & 2092 & - & 215 \\
MCBBS \cite{liu2021leveraging} & \xmark & VGG16 & Q, S & 64.8 & Arria GX1150 & 242 & - & 1785 & 605 & 2704 & 54 & - \\
\midrule
StreamSVD \cite{yu2021streamsvd} & \cmark (partial) & ResNet-18 & Q, TD & 68.4 & XC7Z045 & 125 & - & 752 & - & 576 & - & 34 \\
HPIPE \cite{hall2020hpipe} & \cmark & ResNet-50 & Q, S & 71.9 & S10 2800 & 580 & - & 11278 & 1064 & 10044 & 909 & 4550 \\
FCMP \cite{petrica2020memory} & \cmark & ResNet-50 & Q & 67.3 & Alveo U250 & 195 & 109 & 3870 & 1027 & 1611 & 526 & 2703 \\
\midrule
Ours & \cmark & ResNet-18 & Q, TD & 69.1 & Alveo U250 & 200 & 0 & 3564 & 1550 & 6394 & 1041 & 1138 \\
Ours & \cmark & RepVGG-A0 & Q, TD & 71.5 & Alveo U250 & 200 & 0 & 3550 & 1555 & 5652 & 1162 & 1288 \\
\bottomrule
\end{tabular}
\label{tab:compare}
\end{table*}

\section{Experiments}
\label{sec:exp}
Our experiment is carried out on a server using an Nvidia GTX 1080 Ti GPU for accuracy queries and final model fine-tuning. The accelerator is evaluated using Vivado 2020.1, targeting the AMD Alveo U250 device. 
\subsection{Benchmarks}
As a case study, we focused on the ImageNet dataset and evaluated two state-of-the-art models, ResNet-18 and RepVGG-A0 \cite{ding2021repvgg}. ResNet-18 features representative residual block designs, while RepVGG-A0 is the latest model from the VGG family. Both models were quantized to 8-bit Block Floating Point (BFP) format \cite{banner2019post}. Table~\ref{tab:compression_summary} summarizes the results of our investigation. Our proposed method successfully produced a model with a significantly reduced number of parameters while maintaining an accuracy loss of less than 0.4pp compared to a model that uses 8-bit BFP for both activations and weights.

\begin{table}[h]
    \setlength\tabcolsep{3.5pt}
    \centering
    \caption{Compression results on two CNN benchmarks trained on ImageNet dataset. BitOPs are counted as $2\times W\times A$ MACs. Models have been fine-tuned after the compression. Throughput is on Alveo U250 with a batch size of 1. }
    \begin{tabular}{@{}ccccc@{}}
    \toprule
         ~ & Precision & Float32 & BFP-W8A8 & BFP-W8A8 \\
         ~ & Decomposed & \xmark & \xmark   & \cmark \\
    \midrule     
    \multirow{4}*{ResNet-18} &     Top-1 Accuracy (\%) & 69.7 & 69.3 & 69.1 \\
    ~&     Memory Size (Mb) & 374 & 94 & 35 \\
    ~&      BitOPs (G) & 3715 & 232 & 97 \\
    ~&      Throughput (FPS) & N/A & 702 & 1041\\
        \midrule     
    \multirow{4}*{RepVGG-A0} &     Top-1 Accuracy (\%) & 72.4 & 71.9 & 71.5 \\
    ~&     Memory Size (Mb) & 266 & 67 & 35\\
    ~&     BitOPs (G) & 2789 & 174 & 87 \\
    ~&     Throughput (FPS)  & N/A & 864 & 1162 \\
    \bottomrule
    \end{tabular}
    \label{tab:compression_summary}
\end{table}

\subsection{Performance Results}
We used the proposed approach to generate designs and compared their performance against state-of-the-art work that targets the same task i.e. ImageNet classification using similar type models. Even though a direct comparison with those approaches is not possible as each one utilises a different model and device, it is useful to position the work against the state-of-the-art on the task of ImageNet classification within the space of achieved throughput and accuracy. The results are shown in Table~\ref{tab:compare}. The produced designs by the proposed approach outperform significantly, non-dataflow designs \cite{gong2022n3h, sun2022film, liu2021leveraging} with respect to both peak FPS, and latency (inverse of FPS for batch size 1). In terms of FPS/DSP, our designs outperform non-dataflow designs between 1.73$\times$ and 10.29$\times$.

With respect to dataflow architectures, StreamSVD \cite{yu2021streamsvd} utilises tensor decomposition based on SVD only to compress the weights of the model, which is the closest approach to our work. Their work is a partial dataflow design as device reconfiguration is required to overcome the resource constraint. When compared with them, our work achieves 3.02$\times$ peak FPS/DSP. HPIPE \cite{hall2020hpipe} utilises weight sparsity and we are outperforming them in terms of batch size 1 but not peak performance because their design is clocking at a much higher frequency than ours, 580 MHz versus 200 MHz. FCMP is based on FINN \cite{petrica2020memory} and exploits binary quantization. Because of the binarization, their design utilizes much fewer DSPs than ours but their classification accuracy is lower than ours by 1.8pp on ResNet. Overall, the results show that our Mixed-TD approach can lead to designs with competitive performance with similar task accuracy on ImageNet. 

\begin{figure}[t]
    \centering
    \includegraphics[width=0.48\textwidth]{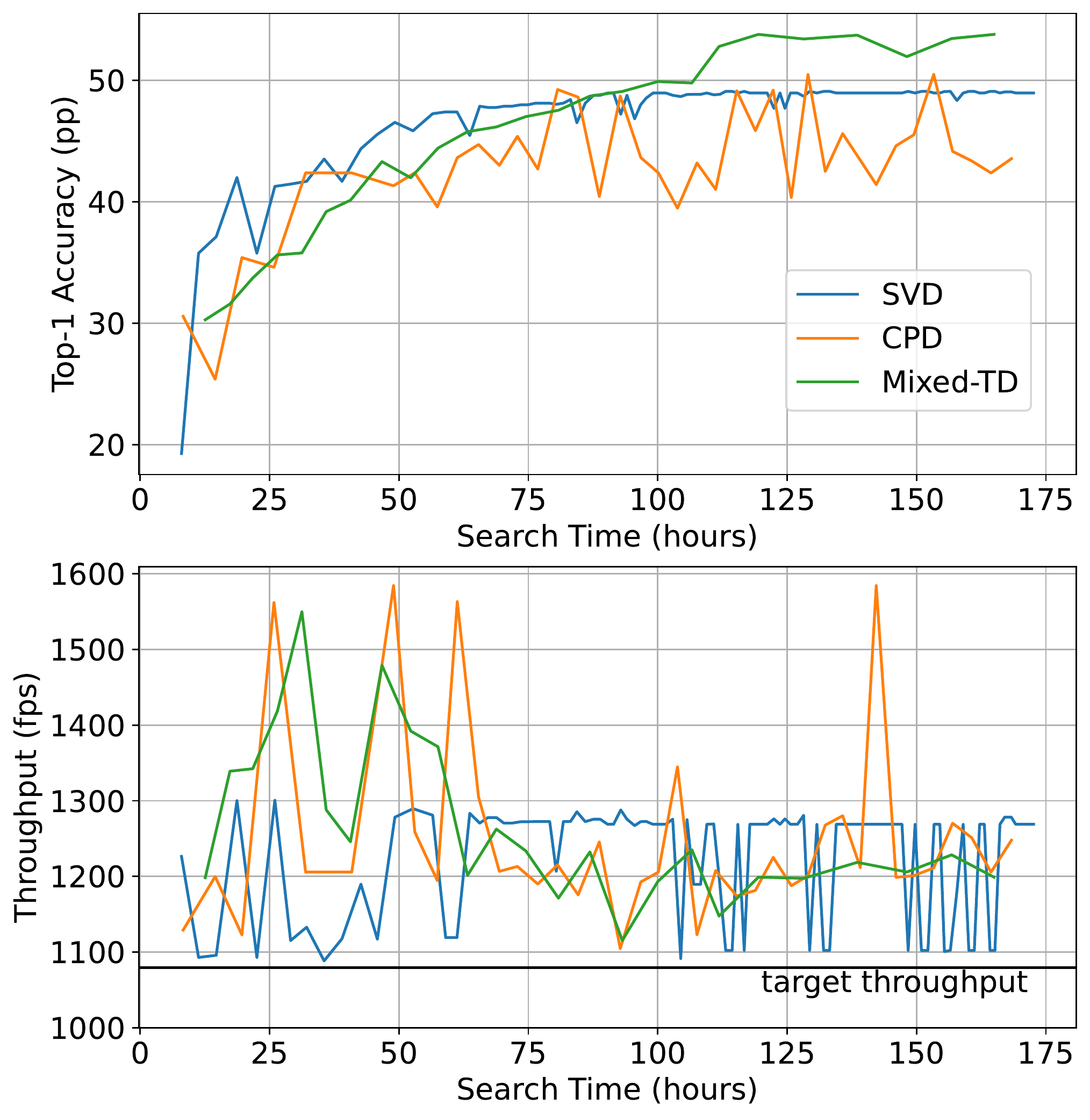}
    \caption{Compare different tensor decomposition methods on ResNet-18. The searching is without the involvement of the performance predictor. Curves do not start from zero points as the initialisation is required to obtain the first generation of valid design points.}
    \label{fig:search_1}
\end{figure}
\begin{figure}[t]
    \centering
    \includegraphics[width=0.48\textwidth]{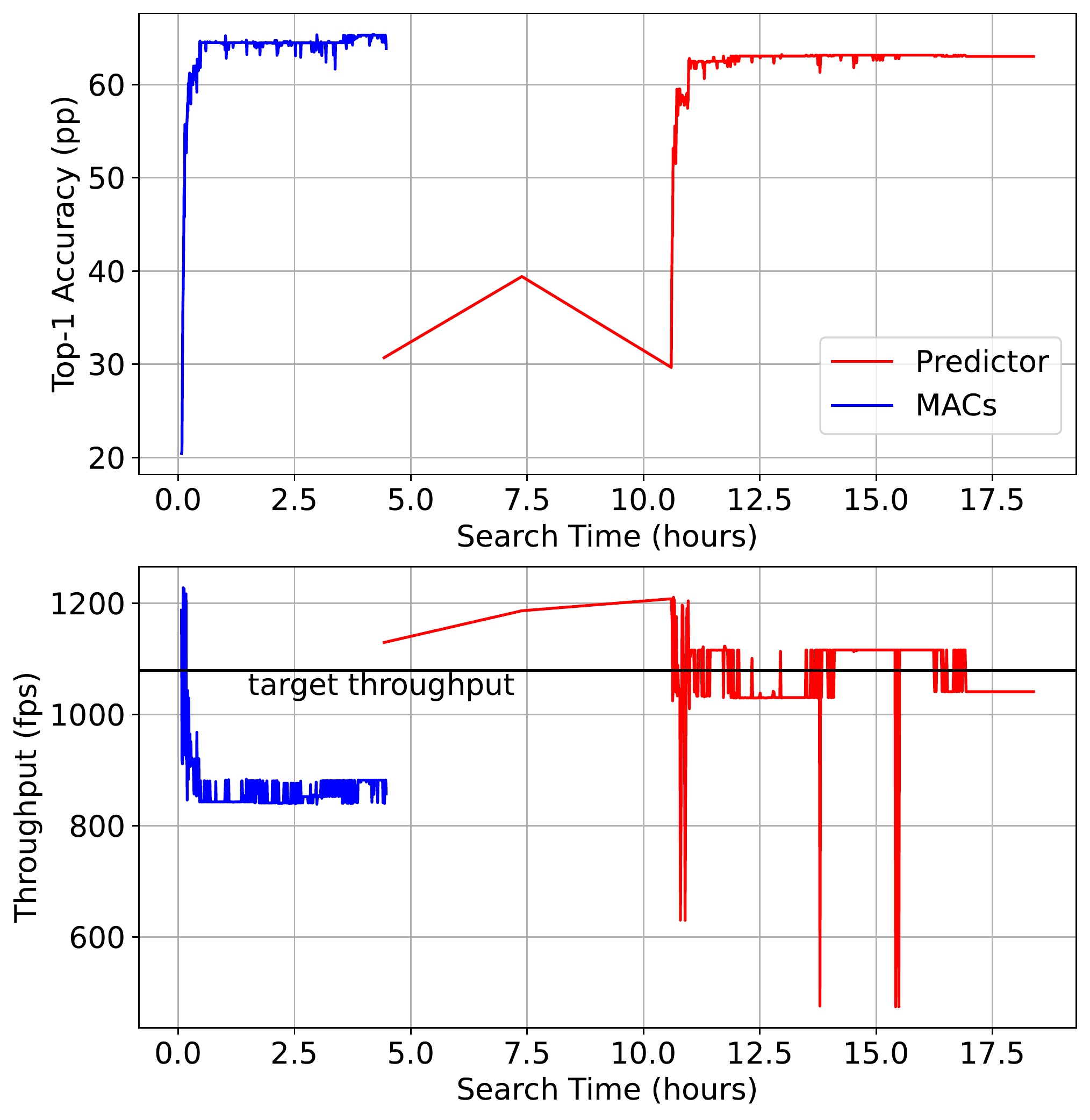}
    \caption{Compare the searching results on ResNet-18, Mixed-TD, when using the predictor to estimate throughput v.s. using MACs for the estimation. The first few steps of the predictor run are slow as the solver \cite{montgomerie2022samo} is launched to gather the training data. Our predictor converges to a design point that better meets the target.}
    \label{fig:search_2}
\end{figure}
\subsection{Ablation Studies}
To better understand the individual contributions of the two main components of the work, mixed tensor decomposition and random-forest-based predictor, ablation studies were conducted. These studies allowed us to analyze the impact of each component on the overall performance of the system.

The top-1 accuracy achieved over time when SVD-only, CPD-only, and the proposed Mixed-TD approaches are used for the decomposition of the weights tensors are illustrated in Fig.~\ref{fig:search_1}. As the Mixed-TD approach explores a larger design space, the benefits in accuracy are only observed after the 90-hour mark.

We investigated the performance benefits of other decomposition methods such as Tensor Train \cite{oseledets2011tensor} and Tensor Ring \cite{zhao2016tensor} into our Mixed-TD algorithm, but we did not observe any further gains to the benchmarks that we investigated.

We have evaluated the performance of our proposed performance predictor for accelerating the design space exploration and compared it to a baseline approach that uses the number of Multiply-Accumulate (MAC) operations to guide the exploration. As shown in Fig.~\ref{fig:search_2}, while the MAC-based approach is $10\times$ faster, it  leads to designs with significantly lower throughput than the target. On the other hand, our performance predictor converges to a design point that better meets the target throughput, resulting in significant time savings compared to a full optimization process for mapping the model to an FPGA. Please note that the full optimization process can only explore 22 designs/hour and it takes more than 150 hours to converge, where with the help of the predictor, we are able to explore 4969 designs/hour and the searching converges with less than 20 hours.

Furthermore, in our experiment, the target throughput is 1079 FPS, and our predictor identifies the design point with 1041 FPS, which is only $3.5\%$ lower. To build the predictor, we chose a random forest due to its fast training time, typically taking only a few minutes, as opposed to other options such as Graph Convolutional Network \cite{dudziak2020brp} which can be relatively slow to train.

\section{Conclusion}
The paper presents a novel method, called Mixed-TD, for fine-grained and layer-specific model compression. This approach addresses the on-chip memory limitations of dataflow architecture accelerators. Mixed-TD achieves substantial weight compression while preserving high accuracy and considering the target hardware. To efficiently navigate the extended design space, we introduced an evolutionary search with a throughput predictor based on a random forest. The paper demonstrates the benefits of tensor decomposition methods in the space of mapping CNN models onto FPGAs, as well as the need for proxies in order to navigate quickly the large design space.

\section*{Acknowledgement}
For the purpose of open access, the author(s) has applied a Creative Commons Attribution (CC BY) license to any Accepted Manuscript version arising.

\bibliographystyle{IEEEtran}
\bibliography{bibliography}

\end{document}